# Anatomy Guided Coronary Artery Segmentation from CCTA Using Spatial Frequency Joint Modeling

Huan Huang[1], Michele Esposito[2], Chen Zhao[1*]

1. Department of Computer Science, Kennesaw State University, Marietta, GA, 30060

2. Department of Cardiology, Medical University of South Carolina, Charleston, SC, USA

**Abstract:**

**Objective:** Accurate coronary artery segmentation from coronary computed tomography angiography is essential for quantitative coronary analysis and clinical decision support. Nevertheless, reliable segmentation remains challenging because of small vessel calibers, complex branching, blurred boundaries, and myocardial interference.

**Methods:** We propose a coronary artery segmentation framework that integrates myocardial anatomical priors, structure aware feature encoding, and three dimensional wavelet inverse wavelet transformations. Myocardial priors and residual attention based feature enhancement are incorporated during encoding to strengthen coronary structure representation. Wavelet inverse wavelet based downsampling and upsampling enable joint spatial frequency modeling and preserve multi scale structural consistency, while a multi scale feature fusion module integrates semantic and geometric information in the decoding stage. The model is trained and evaluated on the public ImageCAS dataset using a 3D overlapping patch based strategy with a 7:1:2 split for training, validation, and testing.

**Results:** Experimental results demonstrate that the proposed method achieves a Dice coefficient of 0.8082, Sensitivity of 0.7946, Precision of 0.8471, and an HD95 of 9.77 mm, outperforming several mainstream segmentation models. Ablation studies further confirm the complementary contributions of individual components.

**Conclusion:** The proposed method enables more stable and consistent coronary artery segmentation under complex geometric conditions, providing reliable segmentation results for subsequent coronary structure analysis tasks.

1. **Introduction:**

Coronary artery disease (CAD) remains one of the leading causes of mortality worldwide, with both its incidence and death rates steadily increasing over the past two decades, thereby imposing a substantial global public health burden [1], [2]. The fundamental pathophysiology of CAD arises from myocardial ischemia caused by luminal narrowing or occlusion of the coronary arteries, resulting in clinical manifestations that range from stable angina to acute myocardial infarction [3], [4]. Coronary computed tomography angiography (CCTA), as a noninvasive imaging modality with excellent three dimensional visualization capability, has become an essential tool for anatomical assessment, stenosis detection, and pre-procedural planning [5], [6]. Despite the high-quality volumetric information CCTA provides, its clinical interpretation remains heavily dependent on experienced radiologists. Tasks such as coronary segmentation, stenosis quantification, and lesion localization typically require slice by slice examination combined with labor intensive manual or semi-automated post processing, leading to considerable time consumption, increased cost, and substantial inter observer variability [7], [8]. These limitations are particularly pronounced in low contrast regions, small distal branches, and complex bifurcations, where inconsistent human interpretation elevates the risk of misdiagnosis and missed lesions [9], [10]. Consequently, developing an automated coronary segmentation method that achieves high accuracy, strong robustness, and substantial reduction of manual effort holds significant clinical value.

In recent years, deep learning has emerged as the predominant methodological paradigm for automatic coronary artery segmentation in CCTA, with convolutional neural networks (CNNs) built upon U-Net and its three-dimensional variants being the most widely adopted [11], [12]. Song et al. proposed a feature fusion and correction framework based on 3D U-Net, incorporating dense blocks and three-dimensional residual correction modules to enhance high level semantic representation [13]. Hong et al. proposed DAC-UNet, a multi-level feature fusion U-shaped network equipped with a dual attention coordination mechanism to retain fine coronary details, suppress background interference and improve coronary delineation [14]. Serrano Antón et al. systematically compared 2D and 3D U-Net architectures under varying training data scales and employed transfer learning to improve coronary feature extraction in data limited scenarios, achieving notable advantages in detecting small caliber vessels [15]. Despite these advances in local structure modeling, CNNs remain fundamentally limited by restricted receptive fields, making it challenging to capture the long-range trajectories, hierarchical branching patterns, and complex topologies of coronary trees.

To overcome the limitations of CNNs in modeling longr-ange dependencies and complex vascular structures, Transformer architectures have been increasingly adopted in medical image analysis [16], [17]. Their global self-attention enables more effective capture of cross regional semantic relationships, improving structural consistency[18]. In medical image segmentation, representative approaches such as Trans U-Net [19] integrate Transformer modules into the U-Net encoder, while Swin Transformer and its 3D variants [20] employ window based attention and hierarchical representations for efficient multi scale modeling. In coronary imaging, Zhao et al. introduced TMN-Net, a hybrid 2.5D multi branch Transformer framework that enhances multi scale coronary representation through combined 2.5D and 3D contextual encoding [21]. Dong et al. proposed a parallel ViT–CNN encoder with variational fusion to integrate global geometry and local detail, supported by uncertainty modeling for cross domain robustness[22]. Yang et al. developed a channel enhanced Transformer with multi scale attention fusion to improve small vessel and bifurcation delineation [23]. Despite these advances, Transformer models still face substantial computational and memory burdens in high resolution 3D CCTA [24], [25] and lack explicit constraints on high frequency edges, making them prone to local structural drift in stenotic or low contrast regions[26], [27].

Despite the continuous progress of deep learning in structural representation, existing methods still face critical limitations when applied to real CCTA data. Mainstream networks rely on max pooling or strided convolutions for downsampling, operations that suppress high frequency boundaries and cause distal vessels and complex bifurcations to disappear in low contrast regions [28]. Deconvolution and interpolation based upsampling often introduce checkerboard artifacts and boundary shifts during complex geometric reconstruction, undermining vascular shape consistency and affecting downstream stenosis assessment [29]. Moreover, most approaches remain confined to spatial domain modeling and fail to exploit the inherent multi scale separability of CCTA in the frequency domain: high frequency components capture vascular edges and stenotic details, whereas low frequency components encode global anatomy, yet conventional convolutions cannot jointly preserve and coordinate these signals [30]. In addition, the absence of anatomical priors leads to over expansion and false positives in regions where myocardial–vascular boundaries are ambiguous, further increasing the risk of mis segmentation [31].

To address these challenges, this study proposes a coronary artery segmentation framework that integrates myocardial anatomical priors, structure aware feature encoding, and three-dimensional wavelet inverse wavelet transformations. During encoding, myocardial region constraints and a residual attention feature enhancement mechanism is introduced to strengthen the representation of coronary relevant structures. During decoding, multi scale feature fusion combined with joint spatial frequency modeling enables coordinated reconstruction of vascular structures across different scales, thereby improving the stability and consistency of segmentation results under complex vascular geometries. The main contributions are as follows:

1. Anatomical prior guided feature encoding. We propose a myocardial region constrained coronary segmentation framework that leverages expanded myocardial contours as anatomical priors. This guidance steers the encoder to focus on spatial regions highly correlated with coronary trajectories, resulting in improved global morphological consistency.

2. Residual–attention feature enhancement with spatial frequency modeling. A residual feature encoder (RFE) is introduced to strengthen coronary structure related feature representation. In addition, wavelet inverse wavelet based downsampling and upsampling modules are designed to jointly model spatial and frequency domain features, preserving low frequency anatomical contours while enhancing high frequency boundary details.

3. Progressive multi scale feature fusion for robust decoding. A multi scale feature fusion (MSFF) module is developed in the decoder to progressively integrate features from different resolutions, allowing deep semantic information and shallow geometric details to be coordinated within a unified framework, thereby improving segmentation stability in complex coronary geometries.

**2. Method**

**2.1 Overall Architecture**

As illustrated in Fig. 1, we propose a coronary artery segmentation framework that integrates myocardial anatomical priors with a three-dimensional wavelet inverse wavelet transformation, designed to simultaneously enhance fine vessel structural preservation and overall segmentation stability. The model adopts a symmetric U-shaped architecture composed of myocardial prior encoder (MPE), RFE, wavelet driven downsampling modules, inverse wavelet decoder and MSFF. Along the encoding path, myocardial region constraints progressively guide extraction toward potential coronary trajectories. During downsampling, three-dimensional discrete wavelet transforms replace conventional max pooling operations to preserve high frequency boundary information and provide stable multi scale representations from the frequency domain perspective. The decoding path reconstructs spatial resolution through inverse wavelet upsampling without introducing checkerboard artifacts.

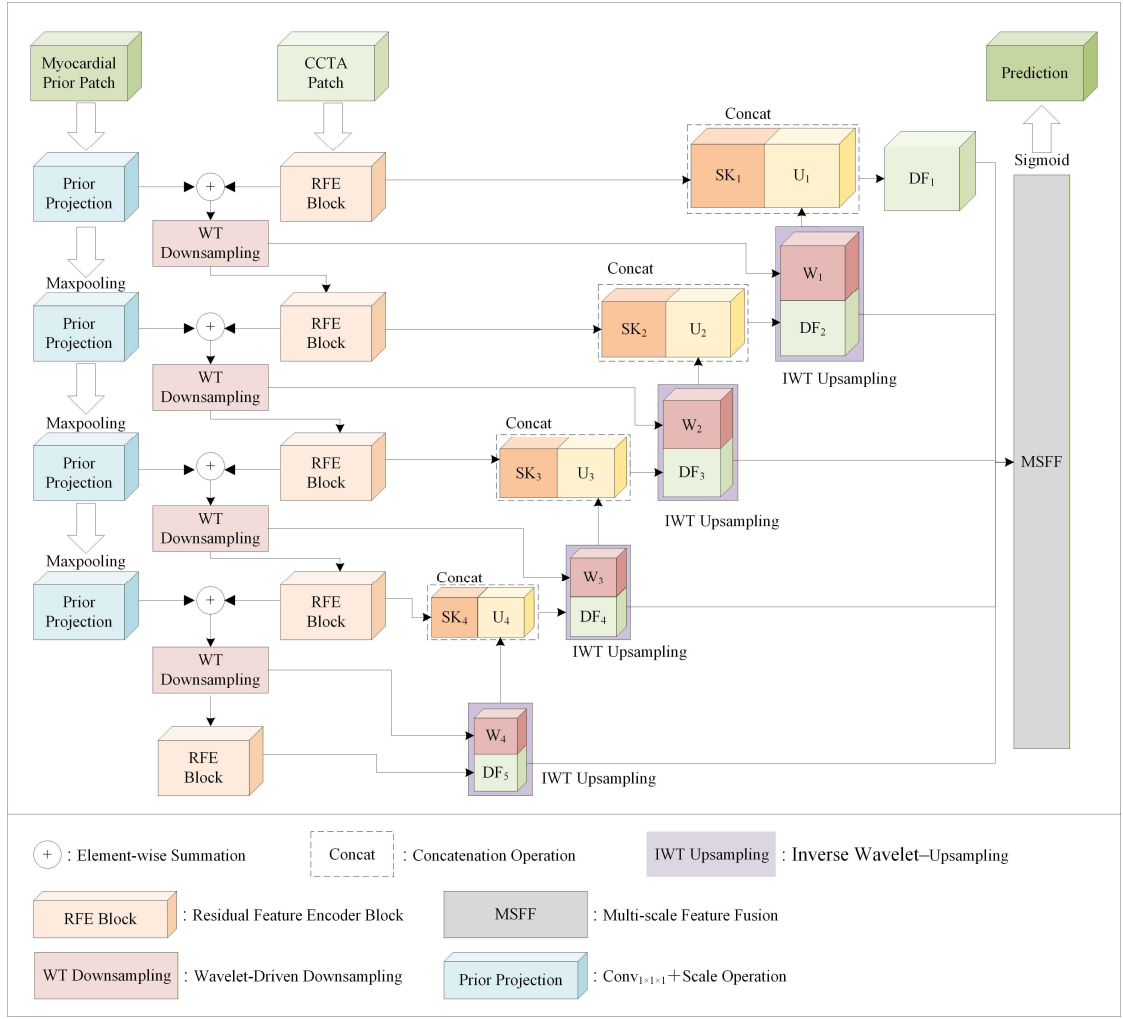

Figure 1. Overall architecture of the proposed coronary artery segmentation network. The encoder integrates myocardial prior projection and residual features encoding with wavelet driven downsampling, generating encoder features $SK_i$ and wavelet subbands $W_i$. In the decoder, inverse wavelet upsampling combines the deeper decoder feature $DF_{i+1}$ with $W_i$ to produce the upsampled feature $U_i$, which is concatenated with $SK_i$ and refined to obtain the decoder output $DF_i$. The multi scale decoder features $\{DF_1, DF_2, DF_3, DF_4\}$ are fused by MSFF to generate the final prediction.

**2.2 Myocardial Prior Encoder**

Coronary arteries are anatomically distributed along the epicardial surface, with their main trunks and branches coursing closely over the myocardium. This spatial relationship provides a natural anatomical priority: constraining the search space to the myocardial region helps delineate plausible vascular pathways, suppress background interference, and improve discrimination in ambiguous areas. Motivated by this clinical characteristic, we first trained a U-Net based myocardial segmentation model using 60 manually annotated CCTA scans and applied it to the full dataset to automatically extract three-dimensional myocardial regions.

Owing to the continuous and closed nature of the myocardium in three-dimensional space, the initial segmentation undergoes connected component filtering followed by slice-wise contour extraction and morphological dilation to generate an expanded myocardial prior covering the epicardial surface. To embed this anatomical prior into multi scale encoding, we introduce a prior projection module that operates before spatial downsampling. Specifically, given the myocardial prior $M_{i-1} \in R^{H \times W \times D}$ at the previous scale, a $1 \times 1 \times 1$ convolution followed by a learnable scaling factor is first applied to adjust the channel dimension and control the injection strength of the prior. The projected priority is then spatially downsampled to match the resolution of the next encoder stage as shown in Eq. 1.

$$M_i = \text{DownSample}(\text{Scale} \cdot \text{Conv}_{1 \times 1 \times 1}(M_{i-1})) \qquad (1)$$

where $Conv_{1 \times 1 \times 1}$ denotes channel adjustment, $DownSample$ represents a max pooling operation, and $Scale$ is a learnable scalar initialized to 0.1. Through four recursive stages, multi scale projected priors $\{M_1, M_2, M_3, M_4\}$ are obtained and fused with the corresponding encoder features.

**2.3 Residual Feature Encoder**

To robustly extract multi scale coronary features, the encoding path employs a residual feature encoder as the fundamental feature extraction unit. As illustrated in Fig. 2, the RFE is built upon residual convolutional blocks. Each residual block consists of two consecutive $3 \times 3 \times 3$ convolutional blocks, where each block is composed of a convolution layer followed by batch normalization and ReLU activation. A skip connection is introduced to aggregate the input feature with the residual mapping, alleviating gradient attenuation while preserving local structural details.

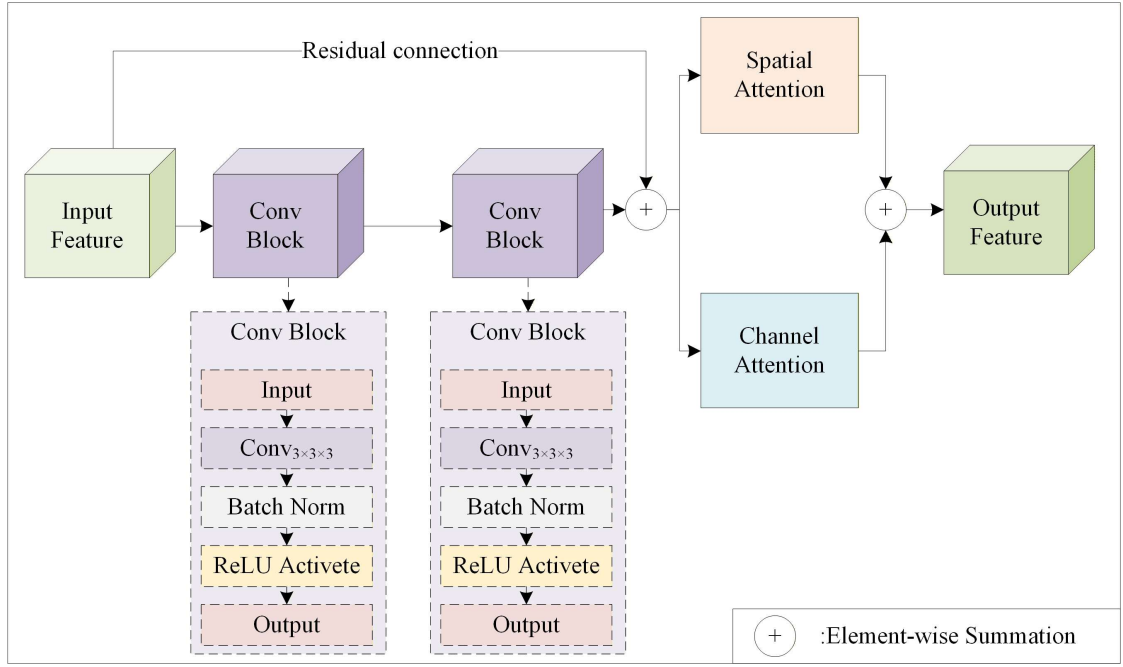

Figure 2. Architecture of the residual feature encoder. The module consists of stacked residual convolutional blocks followed by parallel channel attention and spatial attention branches. Element-wise summation is adopted for residual learning and attention fusion, enabling enhanced representation of coronary artery structures.

Let $X_i$ denote the myocardial guided input feature at scale $i$. The residual convolutional block produces an intermediate representation as defined in Eq. 2

$$R_i = \mathcal{F}(X_i) + X_i \tag{2}$$

where $\mathcal{F}(\cdot)$ denotes the residual mapping implemented by the stacked convolutional blocks. To enhance structure aware feature discrimination, parallel channel attention and spatial attention mechanisms are applied to $R_i$. The channel attention branch models inter channel dependencies by generating a channel-wise weighting vector from globally aggregated features, while the spatial attention branch produces a voxel-wise attention map to emphasize spatially relevant vascular regions. The attention enhanced features are formulated as defined in Eq. 3 and Eq. 4.

$$R_i^{ch} = \sigma\big(g_{ch}(R_i)\big) \odot R_i \tag{3}$$

$$R_i^{sp} = \sigma\big(g_{sp}(R_i)\big) \odot R_i \tag{4}$$

where $g_{ch}(\cdot)$ and $g_{sp}(\cdot)$ denote channel-wise and spatial feature aggregation functions, respectively,

$\sigma(\cdot)$ is the sigmoid activation, and $\odot$ represents element-wise multiplication.

Finally, the outputs of the two attention branches are fused via element-wise summation to generate the encoder output at scale $i$, as shown in Eq. 5.

$$R_i^{out} = R_i^{ch} \oplus R_i^{sp} \tag{5}$$

The resulting feature $R_i^{out}$ serves as a structurally enhanced representation for subsequent downsampling and decoding stages.

**2.4 Wavelet Driven Downsampling Module**

Traditional downsampling operations such as max pooling and strided convolutions tend to discard high frequency boundary information while reducing spatial resolution, causing fine vessels, stenotic regions, and complex bifurcations to gradually degrade in deeper layers. To mitigate this issue, we replace conventional spatial downsampling with a three-dimensional discrete wavelet transform (3D DWT), which explicitly decomposes structural components along different orientations and scales in the frequency domain, thereby preserving both low frequency anatomical contours and high frequency vascular edge details.

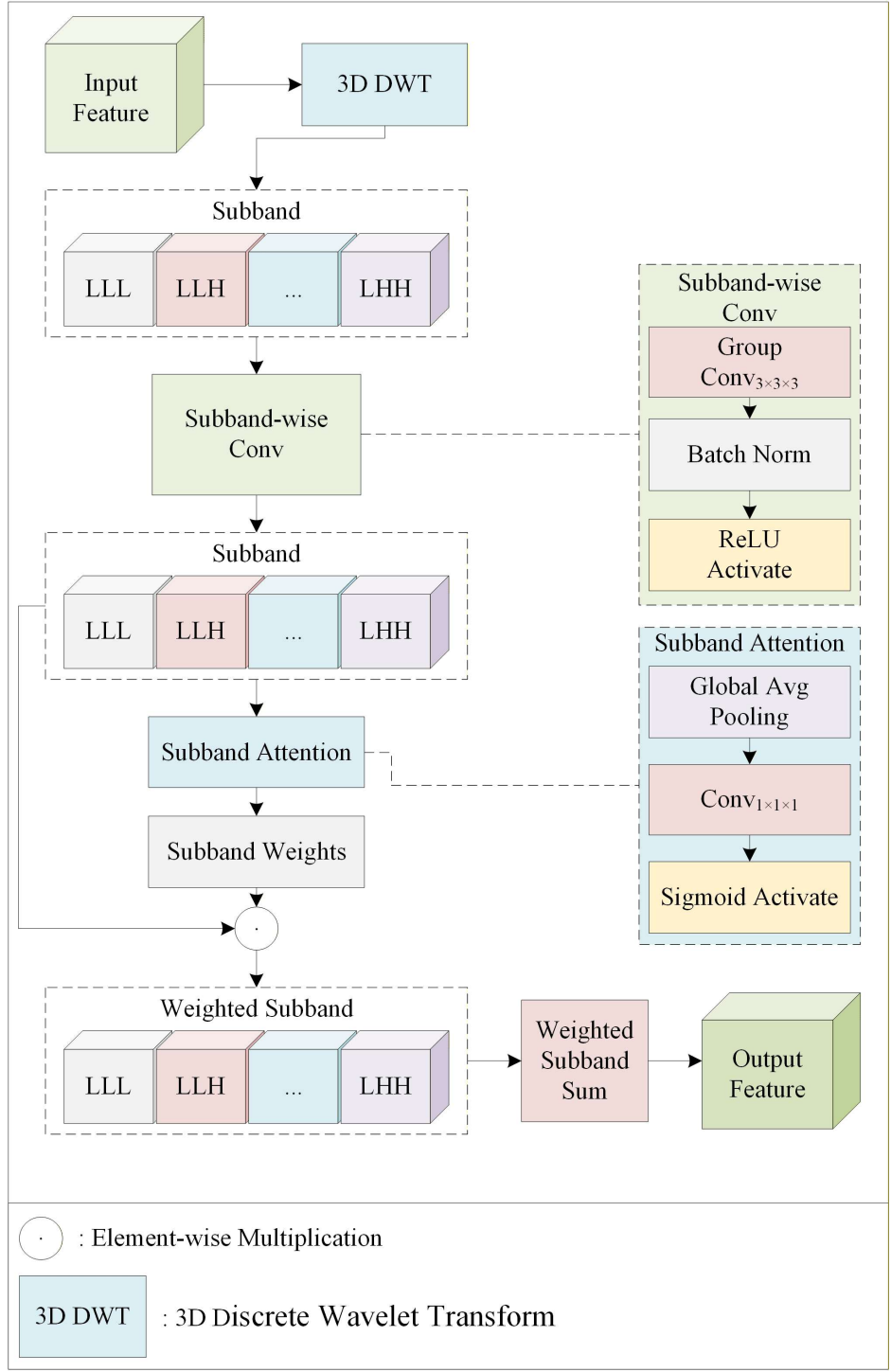

Figure 3. Architecture of the wavelet driven downsampling module. Input features are decomposed into eight directional subbands (LLL-HHH) using 3D DWT, where L and H indicate low and high frequency filtering along each spatial dimension. Subband wise convolution and attention weighting are applied before fusing the subbands into a downsampled feature representation.

As shown in Figure 3, the input feature map be $x \in R^{B \times C \times D \times H \times W}$, where $B, C, D, H,$ and $W$ denote batch size, channel count, and spatial dimensions, respectively.

The wavelet operator $W$, constructed using separable 3D filters, decomposes $x$ into eight directional subbands as defined in Eq. 6.

$$WT(x) = \mathcal{W}(x) \in R^{B \times C \times 8 \times \frac{D}{2} \times \frac{H}{2} \times \frac{W}{2}} \tag{6}$$

Among these subbands, the (LLL) component encodes global low frequency structure, while the remaining high frequency subbands emphasize vessel boundaries, bifurcations, and fine morphological details.

To enhance the expressive capacity of frequency domain features, each subband is processed independently. The eight subbands are first flattened along the channel dimension, as shown in Eq. 7.

$$X_{\text{flat}} = \text{reshape}(WT(x)) \in R^{B \times (8C) \times \frac{D}{2} \times \frac{H}{2} \times \frac{W}{2}} \tag{7}$$

expanding directional subbands into $8C$ channels. A grouped convolution with $g = 8$ is then applied to perform subband specific filtering, as shown in Eq. 8.

$$X_{\text{conv}} = \text{Conv}_{g=8}(X_{\text{flat}}) \tag{8}$$

followed by batch normalization and ReLU activation:

$$X_{\text{enh}} = \text{ReLU}(\text{BN}(X_{\text{conv}})) \tag{9}$$

The enhanced representation is reshaped back into the subband structure:

$$X_{\text{sub}} = \text{reshape}(X_{\text{enh}}) \in R^{B \times C \times 8 \times \frac{D}{2} \times \frac{H}{2} \times \frac{W}{2}} \tag{10}$$

To adaptively select informative directional subbands, a subband attention mechanism is introduced. For the $k$-th subband $X_{\text{sub}}^{(k)}$, its attention weight $a_k$ is computed as Eq. 11.

$$a_k = \sigma\left(\text{Conv}_{1 \times 1 \times 1}\left(\text{Pool}(X_{\text{sub}}^{(k)})\right)\right) \tag{11}$$

where $Pool(\cdot)$ denotes global average pooling and $\sigma$ is a Sigmoid function constraining $a_k \in [0,1]$ to reflect the contribution strength of each subband.

The final frequency domain fusion is obtained as Eq. 12.

$$X_{\text{out}} = \sum_{k=1}^{8} a_k \cdot X_{\text{sub}}^{(k)} \tag{12}$$

Here, $X_{out}$ serves as the primary downsampled feature passed to the encoder pathway, while the original unweighted subbands $WT(x)$ are routed through skip connections to the decoder for inverse wavelet reconstruction and consistent recovery of high frequency structures.

**2.5 Inverse Wavelet–based Decoder**

To avoid the checkerboard artifacts, boundary distortions, and geometric inconsistencies commonly introduced by deconvolution or interpolation based upsampling, the decoder employs a three-dimensional inverse wavelet transform (3D IWT) as the primary spatial reconstruction mechanism. Unlike spatial domain interpolation, IWT leverages the true frequency domain subband information preserved during encoding, ensuring that the upsampling process adheres to frequency consistency principles.

Let the semantic feature at a given decoder scale be $X \in R^{B \times C_{\text{in}} \times D \times H \times W}$, where $B$ denotes batch size, $C_{\text{in}}$ is the channel dimension, and $D, H, W$ are the spatial resolution. To perform IWT, the upsampling module must reconstruct the set of eight directional subbands (LLL, LLH, LHL, …) corresponding to the wavelet decomposition used during encoding. Thus, $1 \times 1 \times 1$ convolution first projects $X$ into predicted subbands as defined in Eq. 13.

$$S_{\text{pred}} = \text{Conv}_{1 \times 1 \times 1}(X) \tag{13}$$

where $S_{\text{pred}} \in R^{B \times C_{\text{out}} \times 8 \times D \times H \times W}$, $C_{\text{out}}$ is the target number of channels after upsampling. During downsampling, the encoder also preserved the corresponding true wavelet subbands $S_{\text{skip}} \in R^{B \times C_{\text{out}} \times 8 \times D \times H \times W}$. To balance semantic prediction with true structural information, a learnable parameter $\alpha \in [0,1]$ is introduced to linearly blend the two sources, as shown in Eq. 14.

$$\tilde{S} = \alpha S_{\text{pred}} + (1 - \alpha) S_{\text{skip}} \tag{14}$$

The fused representation $\tilde{S}$ retains the anatomical fidelity encoded in the true subbands while incorporating semantic cues learned by the decoder, thereby establishing dynamic consistency across high and low frequency components. The inverse wavelet transform then performs upsampling by doubling spatial resolution:

$$Y = \text{IWT}(\tilde{S}, \mathcal{W}_{\text{rec}}) \tag{15}$$

where $\mathcal{W}_{\text{rec}}$ denotes the reconstruction filters paired with the encoder's decomposition filters. The output is a higher resolution feature map $Y \in R^{B \times C_{\text{out}} \times 2D \times 2H \times 2W}$. To compensate for spatial domain fine detail loss and maintain architectural compatibility with U-Net style decoders, the reconstructed tensor $Y$ is concatenated with the encoder's spatial skip feature $E_i$, as shown in Eq. 16.

$$U_i = \text{Concat}(Y, E_i) \tag{16}$$

which is subsequently fed into the decoder convolutional block for joint semantic–geometric reconstruction.

This hybrid decoding strategy combines the artifact free, frequency consistent reconstruction of IWT with the detail preserving spatial skip pathway, enabling stable and accurate recovery of both global coronary topology and fine local vascular geometry.

### 2.6 Multi scale Feature Fusion

Coronary arteries in CCTA exhibit substantial scale variability: large proximal arteries such as the left main and LAD present clear anatomical continuity, whereas distal branches often show fluctuations in size, contrast, and structural coherence. Relying on a single scale representation in the decoding stage is insufficient to simultaneously preserve global vascular topology and fine local geometry. To address this limitation, the proposed framework incorporates a progressive MSFF mechanism along the decoder pathway, enabling consistent reconstruction across different spatial resolutions.

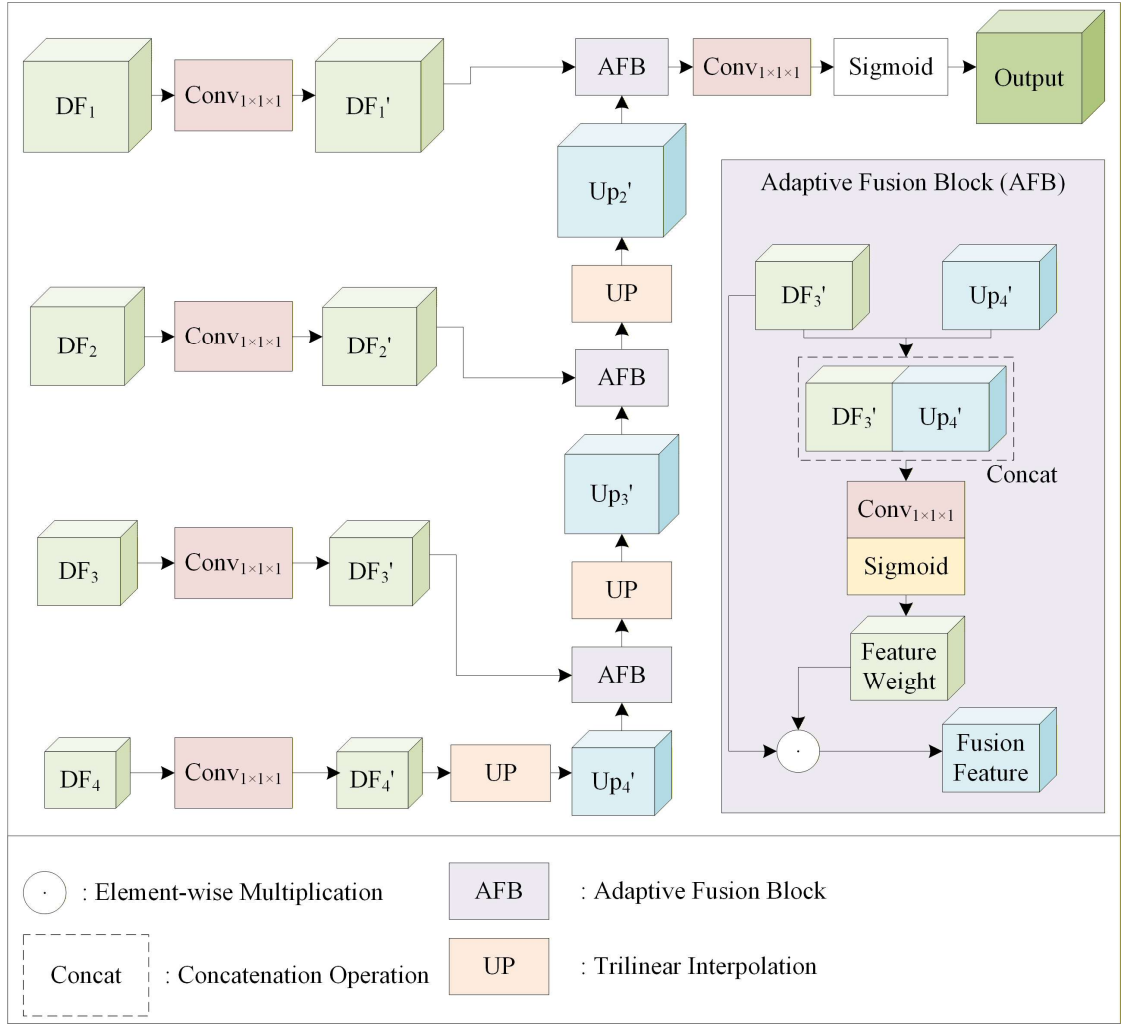

Figure 4. Architecture of the multi scale fusion module. Decoder features at different resolutions are channel aligned and progressively fused in a top-down manner through trilinear upsampling, feature concatenation, and adaptive attention-based weighting, producing the final prediction feature.

As shown in Figure 4, during decoding, the network generates feature maps at four resolution levels, denoted as $\{DF\}_{i=4}^{1}$, with channel dimensions defined by $C_4 = 8C$, $C_3 = 4C$, $C_2 = 2C$, and $C_1 = C$, where $C$ is the base channel width. Each feature map $DF_i \in R^{B \times C_i \times D_i \times H_i \times W_i}$ corresponds to a distinct spatial resolution. To facilitate cross scale fusion, the MSFF module aligns all decoder features to a unified channel dimension of 8 prior to progressive integration, as formulated in Eq. 17.

$$DF_i^{(1)} = \text{Conv}_{1 \times 1 \times 1}(DF_i) \qquad (17)$$

Fusion begins at the deepest scale. The compressed feature $DF_4^{(1)}$ is first upsampled to match the spatial dimensions of $DF_3^{(1)}$:

$$\widehat{DF_4} = \mathrm{Up}(DF_4^{(1)}) \tag{18}$$

where $\mathrm{Up}(\cdot)$ denotes trilinear interpolation. The two features are concatenated along the channel dimension:

$$z_3 = \mathrm{Concat}(\widehat{DF_4}, DF_3^{(1)}) \tag{19}$$

To adaptively model the importance of cross scale information during progressive fusion, an adaptive fusion block is employed to generate fusion weights from the concatenated features, implemented using a $1 \times 1 \times 1$ convolution followed by a Sigmoid activation:

$$w_3 = \sigma(\mathrm{Conv}_{1\times1\times1}(z_3)) \tag{20}$$

The fused representation at scale 3 is obtained as Eq. 21.

$$y_3 = \mathrm{Conv}_{3\times3\times3}(w_3 \odot DF_3^{(1)}) \tag{21}$$

where $\odot$ denotes element-wise weighting. The same recursive fusion is applied to scales 2 and 1, formulated in Eqs. (22)–(25).

$$w_2 = \sigma\left(\mathrm{Conv}_{1\times1\times1}\left(\mathrm{Concat}(\mathrm{Up}(y_3), DF_2^{(1)})\right)\right) \tag{22}$$

$$y_2 = \mathrm{Conv}_{3\times3\times3}(w_2 \odot DF_2^{(1)}) \tag{23}$$

$$w_1 = \sigma\left(\mathrm{Conv}_{1\times1\times1}\left(\mathrm{Concat}(\mathrm{Up}(y_2), DF_1^{(1)})\right)\right) \tag{24}$$

$$y_1 = \mathrm{Conv}_{3\times3\times3}(w_1 \odot DF_1^{(1)}) \tag{25}$$

Finally, the fused shallow scale feature is mapped to a single channel prediction via:

$$P = \mathrm{Conv}_{1\times1\times1}(y_1) \tag{26}$$

Through progressive aggregation of deep semantic representations and shallow geometric features, the MSFF module enables effective cross scale information integration at the prediction stage, contributing to improved structural consistency and boundary delineation in the final segmentation results.

**3. Results**

## 3.1 Experimental Settings

The proposed method was trained and evaluated on the publicly available large scale coronary CT angiography dataset ImageCAS [32]. This dataset contains 1,000 clinical CCTA volumes, each with an in-plane resolution of $512 \times 512$ and $206 - 275$ axial slices. The in-plane pixel spacing ranges from $0.29 - 0.43$ mm, while the slice thickness varies between 0.25–0.45 mm. The cohort covers patients aged $46 - 78$ years. Both left and right coronary trees were independently annotated by two experienced radiologists and subsequently reconciled through cross review to ensure the reliability and consistency of the reference standard.

Prior to training, all volumes were intensity normalized. A 3D patch-based training strategy was employed, where each volume was partitioned into overlapping patches of size $128 \times 160 \times 160$. Adjacent patches shared an overlap of 20 voxels to alleviate spatial discontinuities introduced at patch borders. Random horizontal flipping was used for data augmentation to improve spatial generalization.

All experiments were implemented in PyTorch and conducted on an NVIDIA RTX 4090 GPU. The Adam optimizer was adopted with an initial learning rate of $0.002$, which was progressively decayed using a cosine annealing schedule. Training was performed for 200 epochs with an early stopping criterion of $patience = 15$ to prevent overfitting. The dataset was split into training (700 cases), validation (100 cases), and testing (200 cases) following a 7:1:2 ratio, where the validation set was used for model selection and the held-out test set for final evaluation to ensure independence and reproducibility.

The loss function was defined as a weighted sum of Dice loss $L_{dice}$ and Binary Cross Entropy $L_{bce}$ loss to jointly optimize region overlap and voxel-wise classification accuracy:

$$L_{\text{total}} = \lambda L_{\text{dice}} + (1 - \lambda) L_{\text{bce}} \qquad (27)$$

with the balancing coefficient set to $\lambda = 0.5$. Model performance was evaluated using the Dice similarity coefficient (DSC), Precision, Recall, and the 95th percentile Hausdorff Distance (HD95).

## 3.2 Segmentation Performance

To comprehensively evaluate the proposed method, five representative three-dimensional medical image segmentation networks were selected as baseline models for comparison: (1) 3D U-Net [33], a classical framework in volumetric segmentation featuring a symmetric encoder–decoder architecture with multi scale feature aggregation; (2) V-Net [34], which incorporates residual convolutional blocks and a voxel-wise Dice loss to enhance spatial continuity modeling for organs and vessels; (3) Attention U-Net [35], which introduces explicit attention gating in skip connections to suppress background interference and emphasize foreground relevant regions; (4) SegResNet [36], a deep residual encoder–decoder architecture known for its stability and effectiveness across multi organ segmentation tasks; (5) U-Net R [37], which employs a Transformer based encoder to strengthen long range dependency modeling and pairs it with a U-Net style decoder to preserve global structural consistency.

To ensure a fair comparison, all baseline models were trained from scratch under identical conditions, including dataset split, patch extraction strategy, data augmentation, optimizer configuration, and training schedule. Model selection was based on validation performance, and all quantitative results were reported on the held-out test set.

Table1. Segmentation results of different methods.

| Model | DSC | Sensitivity | Precision | HD95 (mm) |
| --- | --- | --- | --- | --- |
| 3D U-Net | 0.7774 | 0.7051 | 0.8379 | 13.15 |
| V-Net | 0.7989 | 0.7571 | 0.8528 | 9.82 |
| Attention U-Net | 0.8000 | 0.7776 | 0.8305 | 9.98 |
| SegResNet | 0.7994 | 0.7882 | 0.8171 | 10.35 |
| U-Net R | 0.7261 | 0.7415 | **0.8864** | 15.71 |
| **Ours** | **0.8082** | **0.7946** | 0.8471 | **9.77** |

*DSC: Dice Similarity Coefficient; HD95: 95th percentile Hausdorff Distance (mm). Higher DSC, Sensitivity, and Precision indicate better performance, while lower HD95 indicates more accurate boundary reconstruction.*

Table 1 summarizes the quantitative comparison between the proposed method and five mainstream 3D segmentation models on the test set. Overall, the proposed method achieves the highest DSC value (0.8082) and also attains the best performance in Sensitivity and HD95, indicating a well-balanced tradeoff between region overlap accuracy and boundary error control.

Compared with the classical 3D U-Net, the proposed method improves DSC by approximately 3.1 percentage points and reduces HD95 from 13.15 mm to 9.77 mm, demonstrating clear gains in global structural consistency and geometric accuracy. When compared with V-Net and Attention U-Net, the proposed method consistently maintains superior or competitive performance in both DSC and HD95, suggesting that the adopted feature modeling and decoding strategies contribute to improved segmentation stability under identical training configurations. SegResNet achieves a Sensitivity comparable to that of the proposed method, but exhibits slightly inferior DSC and HD95 values, indicating limitations in overall overlap accuracy and boundary delineation. U-Net R attains a relatively high Precision score; however, it lags other methods in DSC and HD95, suggesting that the incorporation of a Transformer based encoder does not translate into a consistent overall performance gain under the current dataset scale and task setting. Although the proposed method does not achieve the highest Precision, its value remains high (0.8471), and when considered alongside its advantages in DSC and HD95, this reflects a more favorable balance between false positive control and structural completeness.

**3.3 Visual Assessment**

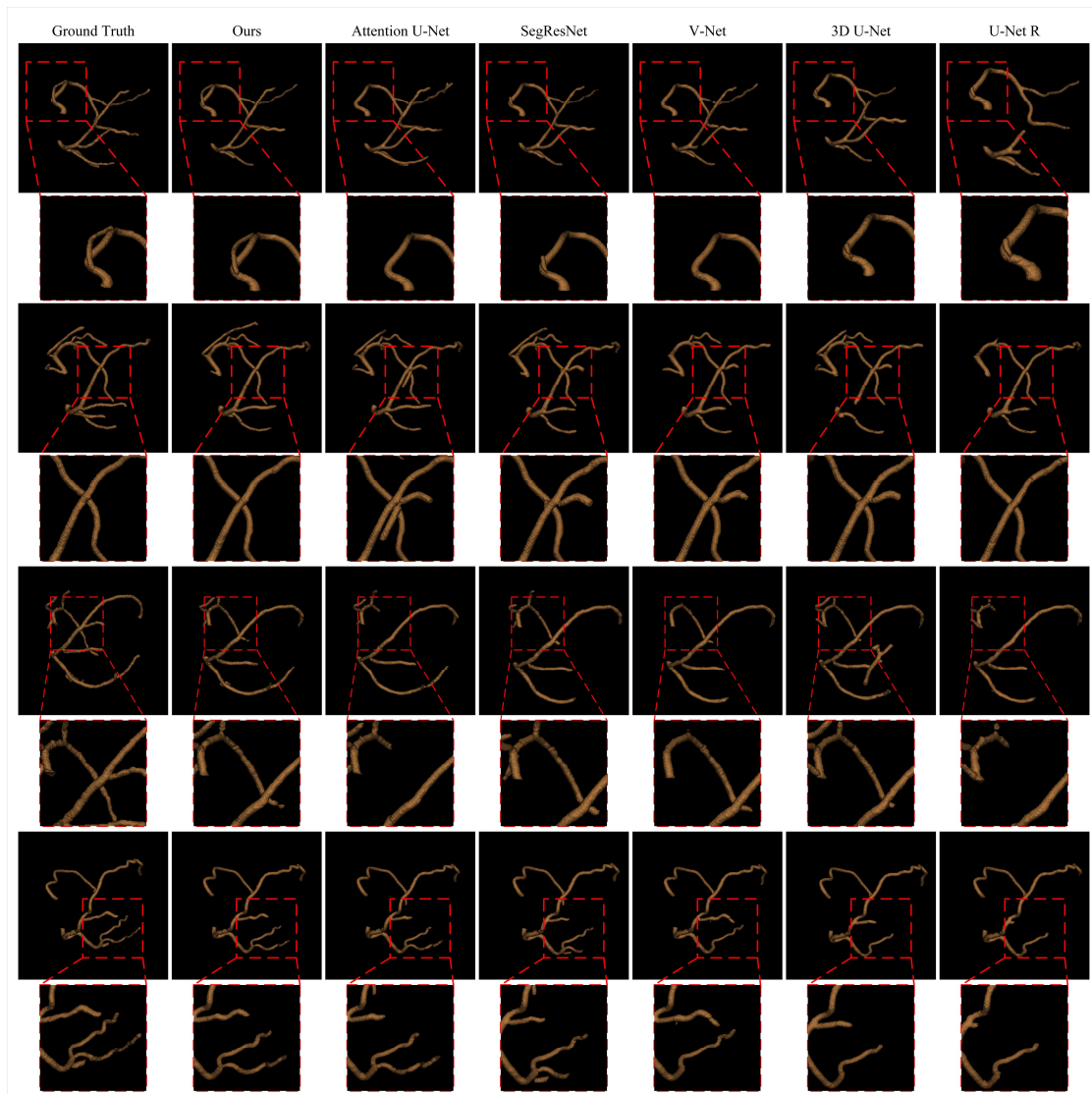

Figure 5. Qualitative visualization of coronary artery segmentation results obtained by different methods. Red boxes highlight representative regions of interest.

To complement the quantitative evaluation, a visual assessment was conducted on several representative cases from the test set, with qualitative comparisons against multiple baseline methods. As illustrated in Fig. 5, the red boxes highlight representative regions of interest, including continuous segments along the main coronary trunks and anatomically complex areas near vessel bifurcations. In these regions, the proposed method is able to recover coronary structures more completely, with segmentation results exhibiting good spatial continuity and overall morphological consistency with manual annotations.

In regions characterized by rapid changes in vessel caliber and complex branching patterns, the proposed method produces vessel contours that appear visually smoother and geometrically more stable, with relatively small deviations from the reference annotations. In contrast, the comparison methods tend to suffer from local discontinuities or unstable boundaries in the same regions. The corresponding three-dimensional coronary reconstructions further demonstrate that the proposed method better preserves the coherence between the main trunk and major branches, resulting in a reconstructed vascular tree that more closely resembles the true anatomical structure.

**4 Discussion**

**4.1 Ablation Experiments**

To systematically analyze the contribution of each key component in the proposed network to coronary artery segmentation performance, ablation experiments were conducted on the ImageCAS test set. All ablation models were trained and evaluated under identical conditions, including data split, training strategy, optimizer configuration, and evaluation metrics, to ensure fair and comparable results. A standard 3D U-Net style architecture was adopted as the baseline model, upon which the MPE (E1), RFE (E2), MSFF (E3), and the wavelet based downsampling and inverse wavelet upsampling modules (WT/IWT, E4) were progressively incorporated to assess the individual and combined effects of different architectural designs.

Table2. Segmentation results of ablation experiments

| Model | MPE | RFE | MSFF | WT/IWT | DSC | Sensitivity | Precision | HD95 |
|---|---|---|---|---|---|---|---|---|
| Baseline | × | × | × | × | 0.7774 | 0.7051 | 0.8379 | 13.15 |
| E1 | √ | | | | 0.7832 | 0.7209 | 0.8315 | 12.82 |
| E2 | | √ | | | 0.7910 | 0.7342 | 0.8424 | 10.65 |
| E3 | | | √ | | 0.7865 | 0.7303 | 0.8314 | 11. |
| E4 | | | | √ | 0.8001 | 0.7669 | 0.8516 | 10.82 |
| Full model | √ | √ | √ | √ | 0.8082 | 0.7946 | 0.8471 | 9.77 |

*DSC: Dice Similarity Coefficient; HD95: 95th percentile Hausdorff Distance (mm). Higher DSC,*

*Sensitivity, and Precision indicate better performance, while lower HD95 indicates more accurate boundary reconstruction*

As shown in Table 2, the performance gains introduced by different modules vary notably, with the wavelet based downsampling and inverse wavelet upsampling (WT/IWT) contributing the most significant improvement. Compared with the baseline model, introducing WT/IWT alone (E4) yields consistent improvements across all metrics, particularly reducing HD95 from 13.15 mm to 10.82 mm. This substantial decrease indicates that frequency consistent feature compression and reconstruction play a critical role in controlling boundary errors and restoring geometric consistency, outperforming conventional spatial domain downsampling and deconvolution based upsampling.

In contrast, the myocardial prior encoder (E1) and the residual–attention feature enhancement module (E2) primarily improve performance through spatial constraint and feature representation enhancement. Among them, E2 achieves larger gains in DSC and HD95 than E1, suggesting that structure aware feature enhancement in the encoding stage has a more direct impact on overall segmentation quality. Introducing the multi scale feature fusion module alone (E3) provides moderate performance improvements; however, the gains remain limited when effective downsampling and reconstruction mechanisms are absent, indicating that stage scale fusion alone cannot fully compensate for structural information loss incurred during encoding. When all components are jointly integrated into the full model, the network achieves the best performance across all evaluation metrics, with HD95 further reduced to 9.77 mm.

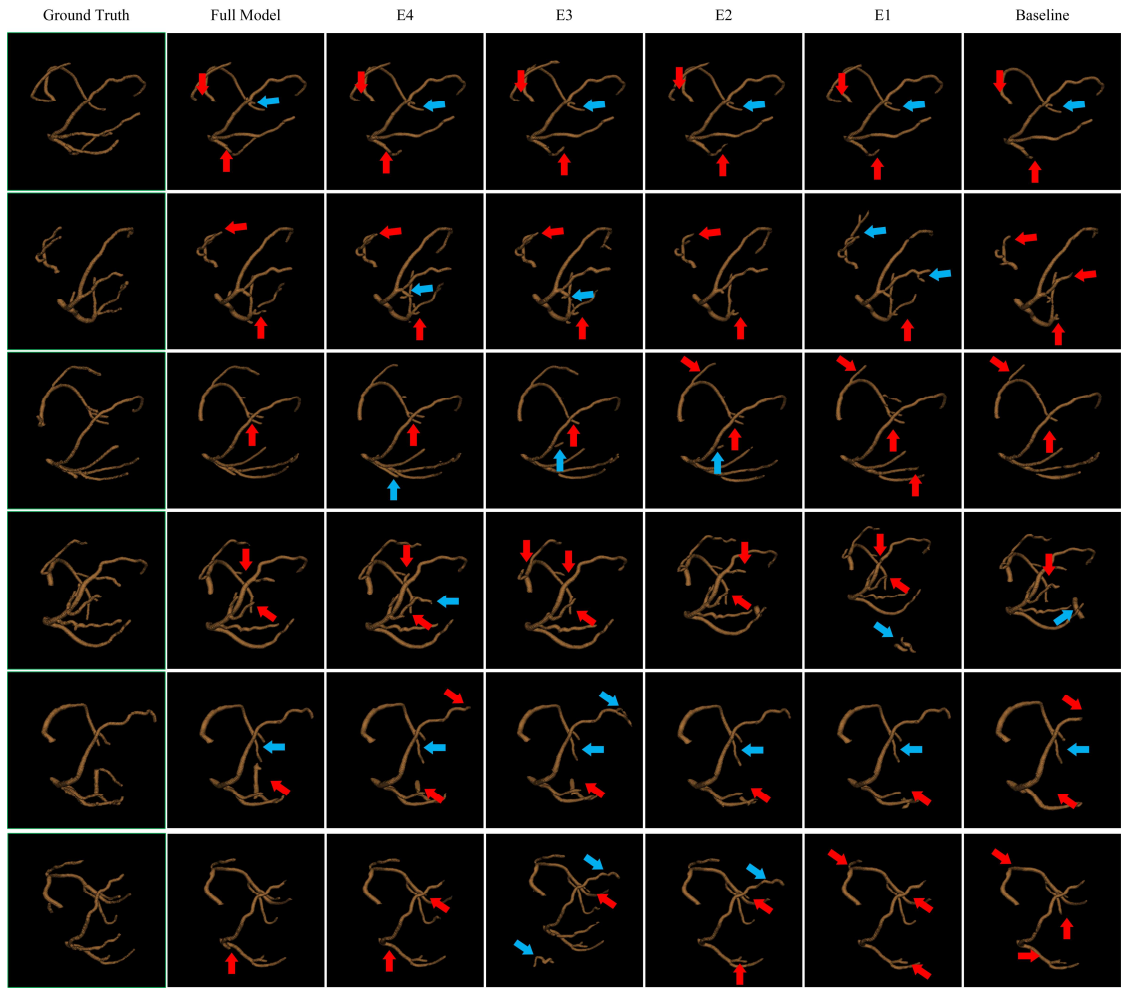

Figure 6. Qualitative visualization of ablation study results. Red and blue arrows indicate representative regions of over segmentation and under segmentation, respectively.

As shown in Figure 6, the ablation study qualitatively demonstrates the complementary roles of frequency domain consistent reconstruction and spatial anatomical constraints. When either component is removed, noticeable over segmentation or under segmentation occurs in representative regions, particularly around complex bifurcation structures, as indicated by the red and blue arrows. By jointly incorporating both components, the full model yields more coherent vascular topology and exhibits improved robustness across challenging anatomical configurations.

**4.2 Limitations and Future Work**

Although the proposed model achieves notable improvements in both quantitative metrics and qualitative visualization, several limitations remain in practical applications. First, due to the inherently three-dimensional nature of CCTA data and the small size and complex morphology of coronary vessels, the complete framework incurs a relatively high computational cost during feature transformation and frequency domain reconstruction. While the WT/IWT mechanism reduces redundancy associated with spatial domain operations, overall inference speed is still constrained by the computational burden of 3D convolutions and grouped convolutions. Further optimization is therefore required to facilitate deployment of resource limited hardware.

Second, the performance of the proposed method depends on the accuracy of myocardial region segmentation. Although myocardial priors provide effective geometric constraints in most cases, local inaccuracies or incomplete myocardial contours may adversely affect the localization of adjacent coronary structures, particularly in cases where myocardial–vascular signal coupling is strong and intrinsic contrast is low. In addition, while the WT/IWT mechanism substantially improves boundary recovery, small distal vessels may still be missed in extremely low contrast, noisy, or heavily calcified regions. This observation suggests that the integration of frequency domain and spatial domain features could be further refined to enhance robustness under challenging imaging conditions.

Finally, the present study focuses on anatomical segmentation and does not explicitly incorporate hemodynamic information, vessel wall characteristics, or plaque composition, which are clinically relevant for comprehensive coronary assessment. Future work may explore extending frequency domain feature modeling to functional evaluation tasks or integrating the proposed framework into multi module pipelines for fractional flow reserve estimation or stenosis quantification, thereby enhancing its overall clinical utility in decision making workflows.

## 5 Conclusion

This study proposes a novel coronary artery segmentation framework that integrates myocardial anatomical priors with three-dimensional wavelet inverse wavelet transformations. By introducing myocardial region constraints during encoding and performing joint spatial frequency modeling in the downsampling and upsampling stages, the proposed method achieves improved stability and accuracy in delineating small vessels, stenotic boundaries, and complex bifurcation structures. Experimental results on a public CCTA dataset demonstrate that the proposed approach outperforms several existing 3D segmentation models across multiple evaluation metrics, including DSC, Precision, Recall, and HD95. Qualitative assessments further confirm their advantages in preserving structural continuity, enhancing boundary clarity, and improving robustness under challenging imaging conditions.

## Acknowledgments

This work is partially supported by American Heart Association under award #25AIREA1377168 (PI. Chen Zhao).